\documentclass{llncs}
\usepackage[latin1]{inputenc} 

\usepackage{amssymb}
\usepackage[pdftex]{graphicx}
\usepackage{amsmath}

\usepackage{tabularx}
\usepackage[ruled,vlined,linesnumbered]{algorithm2e}
\usepackage[T1]{fontenc}
\usepackage{color}
\usepackage{enumitem}
\def\eg{{\em e.g.}}

\def\ie{{\em i.e.}}
\newcommand{\Xfig}[1]{Fig.~\ref{#1}}
\newcommand{\Xfigs}[1]{Figs.~\ref{#1}}
\newcommand{\Xtheo}[1]{Th.~\ref{#1}}
\newcommand{\Xtheos}[1]{Ths.~\ref{#1}}
\newcommand{\Xeqn}[1]{Eq.~\ref{#1}}

\newcommand{\Xsec}[1]{Section~\ref{#1}}

\newcommand{\ELIMINE}[1]{}

\usepackage{subfigure}

\newcommand{\st}[0]{\; | \;}

\newtheorem{lemme}{Lemma}

\newtheorem{thm}[lemme]{Theorem}

\usepackage{makeidx}  
\begin{document}
\title{New characterizations of minimum spanning trees and of saliency
  maps based on quasi-flat zones\thanks{This work received funding
    from ANR (contract ANR-2010-BLAN-0205-03), CAPES/PVE (grant
    064965/2014-01), and CAPES/COFECUB (grant 592/08).}}
\titlerunning{New characterizations of MST}  
%
\author{Jean Cousty$^1$, Laurent Najman$^1$, Yukiko Kenmochi$^1$, Silvio Guimar\~aes$^{1,2}$}
\authorrunning{Jean Cousty et al.} 
\institute{Universit\'e Paris-Est, LIGM, A3SI, ESIEE Paris, CNRS\ELIMINE{Laboratoire d'Informatique Gaspard
  Monge}\\
\email{ \{j.cousty, l.najman, y.kenmochi, s.guimaraes\}@esiee.fr},
\and
PUC Minas - ICEI - DCC - VIPLAB
}
\maketitle              
\vspace{-0.5cm}
\begin{abstract}We study three representations of
  hierarchies of partitions: dendrograms (direct representations),
  saliency maps, and minimum spanning trees. We provide a new
  bijection between saliency maps and hierarchies based on quasi-flat
  zones as used in image processing and characterize saliency maps and
  minimum spanning trees as solutions to constrained minimization
  problems where the constraint is quasi-flat zones preservation. In
  practice, these results form a toolkit for new hierarchical methods
  where one can choose the most convenient representation. They also
  invite us to process non-image data with morphological hierarchies.
\end{abstract}
\section{Introduction}
\label{sec:prevworks}
\begin{figure}[htb]
    \begin{center}
      \includegraphics[height=0.3\linewidth]{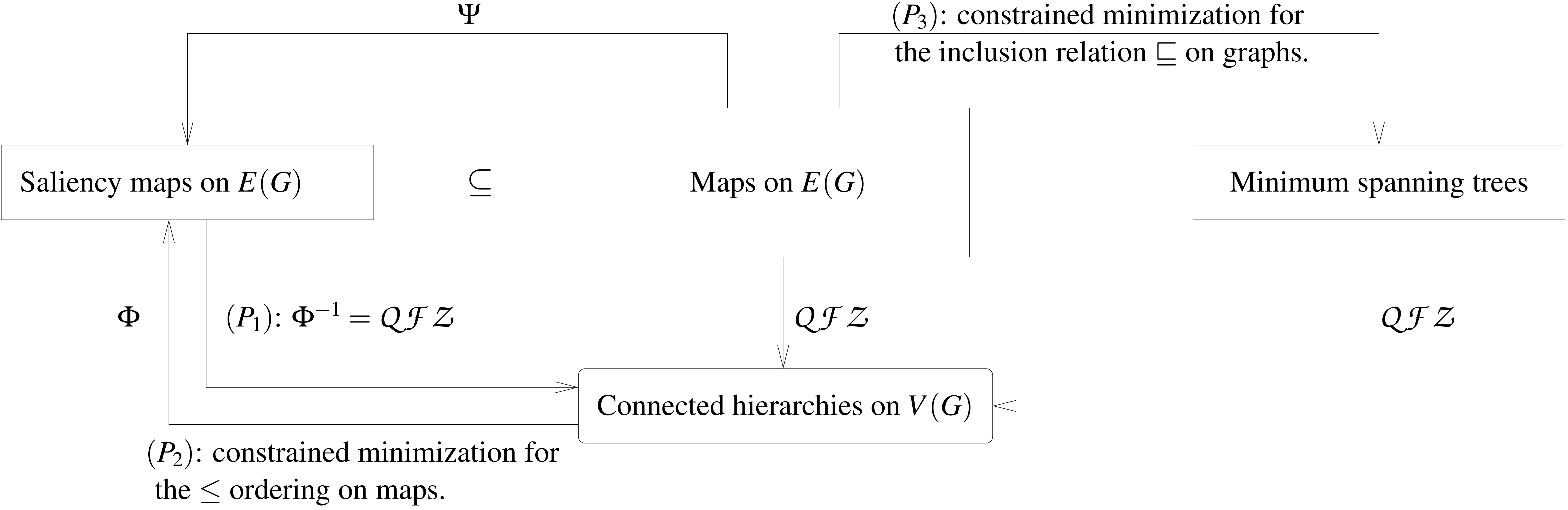}

\caption{\label{fig:diag} A diagram that summarizes the results of
  this article. The solutions to problems~$(P_1)$,~$(P_2)$,
  and~$(P_3)$ are given by
  \Xtheos{thm:SaliencyBijectionConnectedHierarchy},~\ref{thm:SaliencyCharacterization},
  and~\ref{thm:SaliencyMST}, respectively. The constraint involved
  in~$(P_2)$ and~$(P_3)$ is to leave the induced quasi-flat zones
  hierarchy unchanged. In the diagram,~$\mathcal{QFZ}$ stands for
  quasi-flat zones (\Xeqn{eqn:flatZones}), and the symbols~$\Phi$
  and~$\Psi$ stand for the saliency map of a hierarchy
  (\Xeqn{eqn:saliency}) and of a map respectively
  (\Xsec{sec:saliencyCharac}).}
  \end{center}
\end{figure}

Many image segmentation methods look for a partition of the set of
image pixels such that each region of the partition corresponds to an
object of interest in the image. Hierarchical segmentation methods,
instead of providing a unique partition, produce a sequence of nested
partitions at different scales, enabling to describe an object of
interest as a grouping of several objects of interest that appear at
lower scales (see references in
\cite{Najman-and-Cousty-prl2014}). This article deals with a theory of
hierarchical segmentation as used in image processing. More precisely,
we investigate different representations of a hierarchy: by a
dendrogram (direct set representation), by a saliency map (a
characteristic function), and by a minimum spanning tree (a reduced
domain of definition). Our contributions are threefold:
\begin{enumerate}
\item a new bijection theorem between hierarchies and saliency maps
  (\Xtheo{thm:SaliencyBijectionConnectedHierarchy}) that relies on the
  quasi-flat zones hierarchies and that is simpler and more general
  than previous bijection theorem for saliency
  maps;\ELIMINE{\footnote{\Xtheo{thm:SaliencyBijectionConnectedHierarchy}
      extends Th. 13 in \cite{Najman-jmiv2011} both in a simpler way
      (in the sense that quasi-flat zones are considered instead of
      topological watersheds) and in a more general way (in the sense
      that it includes any hierarchy of connected partitions rather
      than hierarchies of edge-segmentations which form a subset of
      hierarchies of connected partitions).}} and
\item a new characterization of the saliency map of a given
    hierarchy as the minimum function for which the quasi-flat zones
    hierarchy is precisely the given hierarchy
    (\Xtheo{thm:SaliencyCharacterization}); and
\item a new characterization of the minimum spanning trees  of a
  given edge-weighted graph as the minimum subgraphs (for inclusion)
  whose quasi-flat zones hierarchies are the same as the one of the
  given graph
  (\Xtheo{thm:SaliencyMST}).
\end{enumerate}
The links established in this article between the maps that weight the
edges of a graph~$G$, the hierarchies on the vertex set~$V(G)$ of~$G$,
the saliency maps on the edge set~$E(G)$ of~$G$, and the minimum spanning
trees for the maps that weight the edges of~$G$ are summarized in the
diagram of \Xfig{fig:diag}.

\begin{figure}[htb]
\begin{center} 	\renewcommand{\tabcolsep}{0.1cm}
\begin{tabular}{cccc} 
\includegraphics[width=0.23\linewidth]{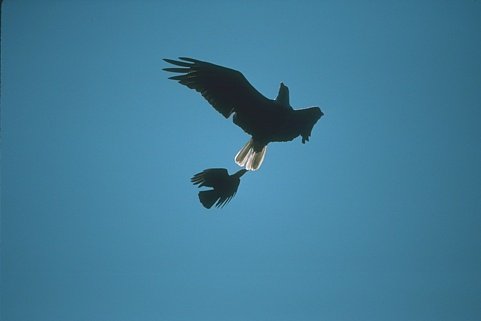} &
\includegraphics[width=0.23\linewidth]{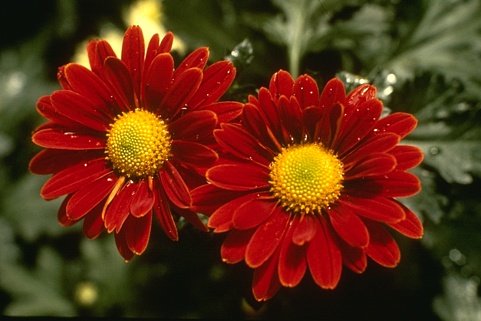} & 
\includegraphics[width=0.23\linewidth]{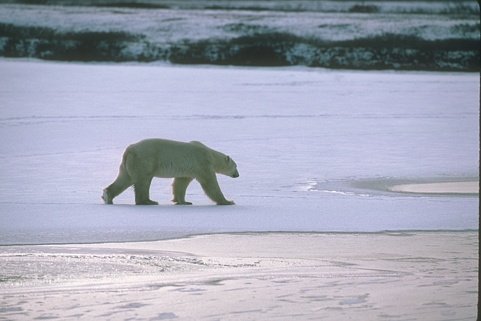} &
\includegraphics[width=0.23\linewidth]{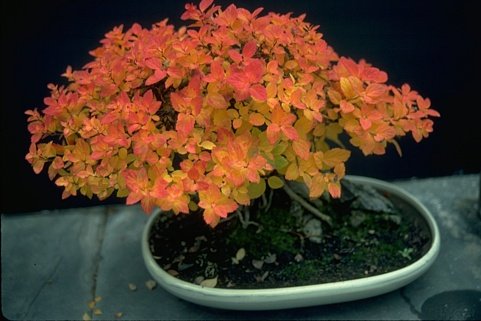} \\ 
{\includegraphics[width=0.23\linewidth]{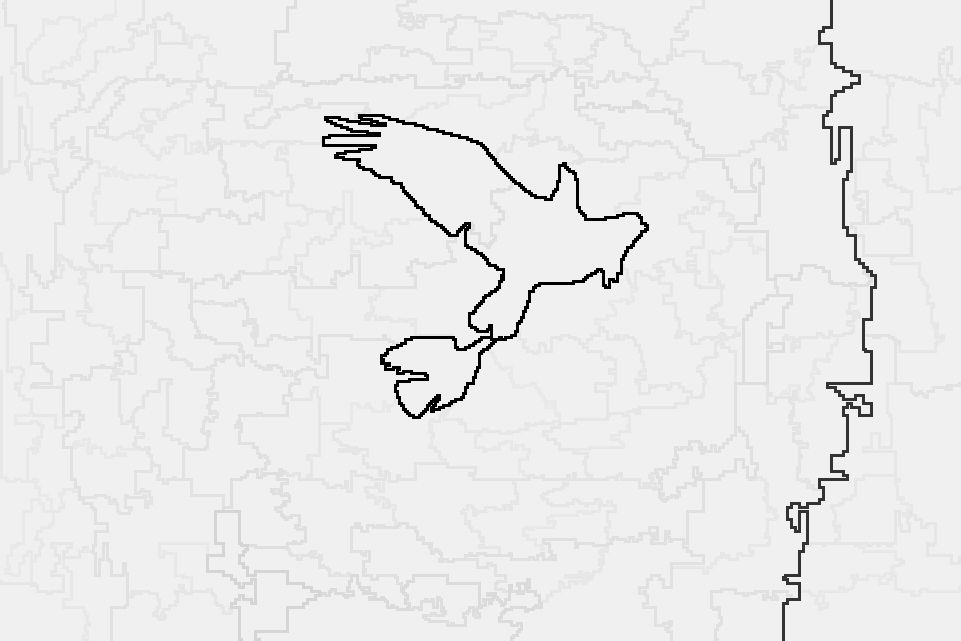}}&
{\includegraphics[width=0.23\linewidth]{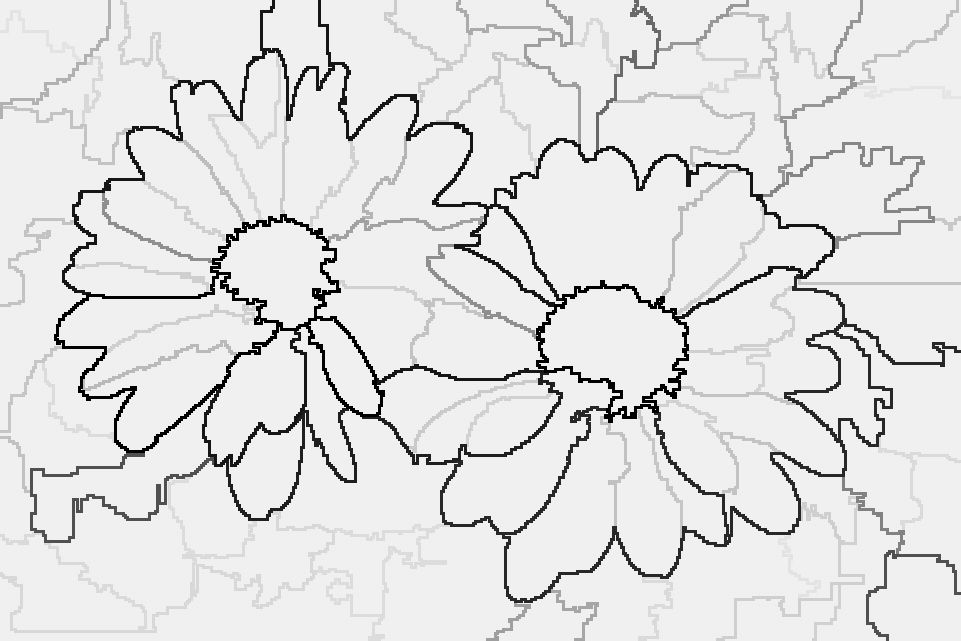}}&
{\includegraphics[width=0.23\linewidth]{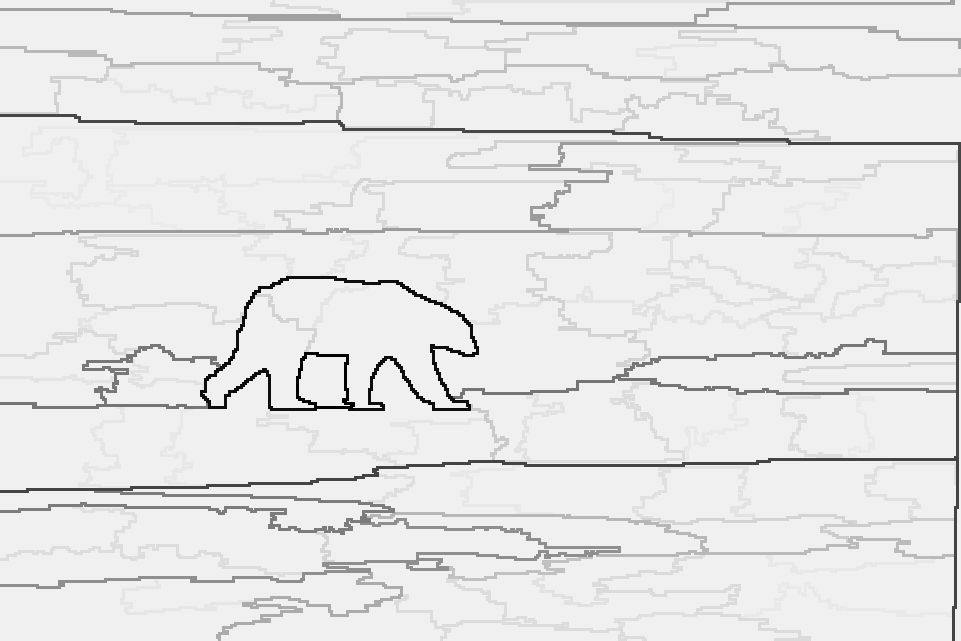}}&
{\includegraphics[width=0.23\linewidth]{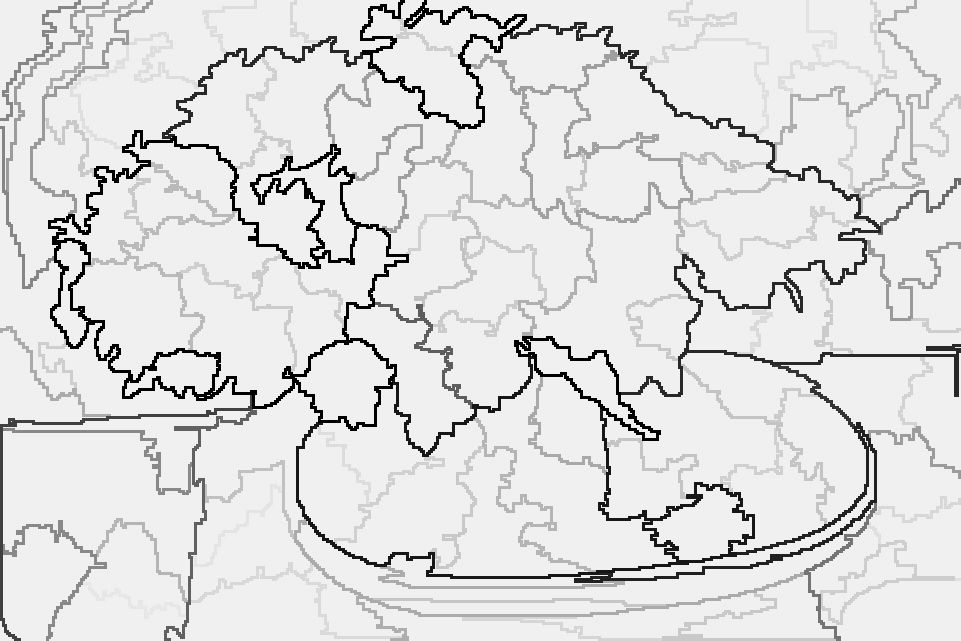}}\\ 
{\includegraphics[width=0.23\linewidth]{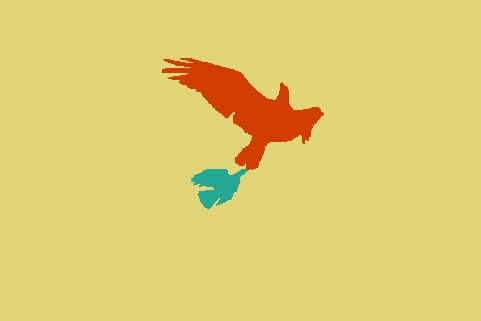}}&
{\includegraphics[width=0.23\linewidth]{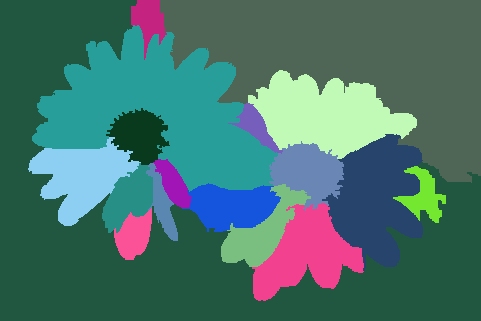}}&
{\includegraphics[width=0.23\linewidth]{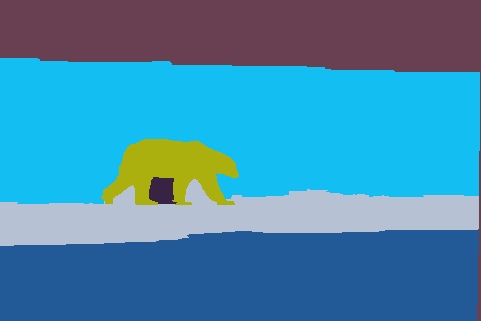}}&
{\includegraphics[width=0.23\linewidth]{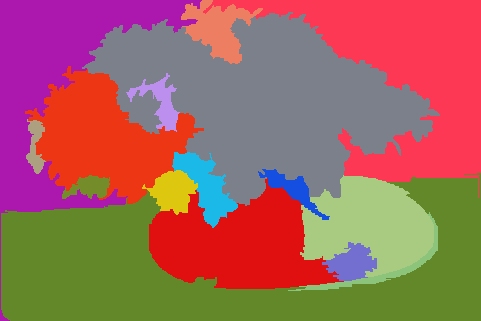}}\\
(a) & (b) & (c) & (d) \\ 
\end{tabular}
\end{center}
\caption{\label{fig:SaillanceSilvio}Top row: some images from the
  Berkeley database~\cite{Arbelaez-et-al-pami2011}. Middle row:
  saliency maps according to \cite{Guimaraes-et-al-ssspr2012}
  developed thanks to the framework of this article.  Bottom row:
  segmentations extracted from the hierarchies with (a) 3, (b) 18, (c)
  6 and (d) 16 regions. }
\end{figure}

One possible application of these results is the design of new
algorithms for computing hierarchies. Indeed, our results allow one to
use indifferently any of the three hierarchical representations.  This
can be useful when a given operation is more efficiently performed
with one representation than with the two others.  Naturally, one
could work directly on the hierarchy (or on its tree-based
representation) and finally compute a saliency map for visualization
purposes. For instance, in
\cite{Guigues-Coquerez-ijcv2006,Kiran-Serra-pr2014}, the authors
efficiently handle directly the tree-based representation of the
hierarchy. Conversely, thanks to
\Xtheo{thm:SaliencyBijectionConnectedHierarchy}, one can work on a
saliency map or, thanks to \Xtheo{thm:SaliencyMST}, on the weights of
a minimum spanning tree and explicitly computes the hierarchy in the
end. In \cite{Cousty-and-Najman-ismm2011,Najman-et-al-ismm2013}, a
resulting saliency map is computed before a possible extraction of the
associated hierarchy of watersheds. In
\cite{Guimaraes-et-al-ssspr2012}, a basic transformation that consists
of modifying one weight on a minimum spanning tree according to some
criterion is considered. The corresponding operation on the equivalent
dendrogram is more difficult to design. When this basic operation is
iterated on every edge of the minimum spanning tree, one transforms a
given hierarchy into another one. An application of this technique to
the observation scale of \cite{Felzenszwalb-ijcv2004} has been
developed in \cite{Guimaraes-et-al-ssspr2012} (see
\Xfig{fig:SaillanceSilvio}).

Another interest of our work is to precise the link between
hierarchical classification \cite{nakache-confais-2004} and
hierarchical image segmentation. In particular, it suggests that
hierarchical image segmentation methods can be used for classification
(the converse being carried out for a long time). Indeed, our work is
deeply related to hierarchical classification, more precisely, to
ultrametric distances, subdominant ultrametrics and single linkage
clusterings. In classification, representation of hierarchies, on
which no connectivity hypothesis is made, are studied since the 60's
(see references in \cite{nakache-confais-2004}). The framework
presented in this article deals with connected hierarchies and a graph
needs to be specified for defining the connectivity of the regions of
the partitions in the hierarchies. The connectivity of regions is the
main difference between what has been done in classification and in
segmentation. Rather than restricting the work done for
classification, the framework studied in this article generalizes
it. Indeed the usual notions of classification are recovered from the
definitions of this article when a complete graph (every two points
are linked by an edge) is considered. For instance, when a complete
graph is considered, a saliency map becomes an ultrametric distance,
which is known to be equivalent to a hierarchy. However,
\Xtheo{thm:SaliencyBijectionConnectedHierarchy} shows that, when the
graph is not complete, we do not need a value for each pair of
elements in order to characterize a hierarchy (as done with an
ultrametric distance) but one value for each edge of the graph is
enough (with a saliency map). Furthermore, when a complete graph is
considered, the hierarchy of quasi-flat zones becomes the one of
single linkage clustering. Hence, \Xtheo{thm:SaliencyMST} allows to
recover and to generalize a well-known relation between the minimum
spanning trees of the complete graph and single linkage clusterings.

\section{Connected hierarchies of partitions}
A {\em partition} of a finite set~$V$ is a set~$\bold{P}$ of nonempty
disjoint subsets of~$V$ whose union is~$V$ (\ie{}, $\forall X,Y \in
\bold{P}$,~$X \cap Y = \emptyset$ if~$X\neq Y$ and $\cup \{X \in
\bold{P}\} = V$). Any element of a partition~$\bold{P}$ of~$V$ is
called a {\em region  of~$\bold{P}$}. If~$x$ is an element
of~$V$, there is a unique region of~$\bold{P}$ that contains~$x$; this
unique region is denoted by~$[\bold{P}]_x$. Given two
partitions~$\bold{P}$ and~$\bold{P}'$ of a set~$V$, we say that
$\bold{P}'$ is a {\em refinement} of~$\bold{P}$ if any region
of~$\bold{P}'$ is included in a region of~$\bold{P}$. A {\em hierarchy
  (on~$V$)} is a sequence~$\mathcal{H} = (\bold{P}_0, \ldots,
\bold{P}_\ell)$ of indexed partitions of~$V$ such
that~$\bold{P}_{i-1}$ is a refinement of~$\bold{P}_i$, for any~$i \in
\{1, \ldots, \ell\}$.  If $\mathcal{H} = (\bold{P}_0, \ldots,
\bold{P}_\ell)$ is a hierarchy, the integer~$\ell$ is called the {\em
  depth of~$\mathcal{H}$}.  A hierarchy~$\mathcal{H} = (\bold{P}_0,
\ldots, \bold{P}_\ell)$ is called complete if~$\bold{P}_\ell = \{V\}$
and if~$\bold{P}_0$ contains every singleton of~$V$ ({\em i.e.},
$\bold{P}_0 = \{\{x\} \st x \in V\}$). The hierarchies considered in
this article are complete.

\begin{figure}[htb]
    \begin{center}
    \begin{tabular*}{1\linewidth}{@{\extracolsep{\fill}}c c c c c}
      \includegraphics[height=0.13\linewidth]{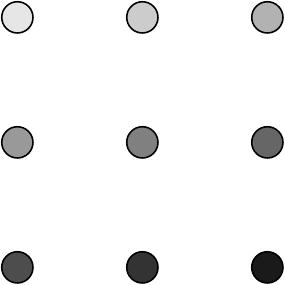}&
      \includegraphics[height=0.13\linewidth]{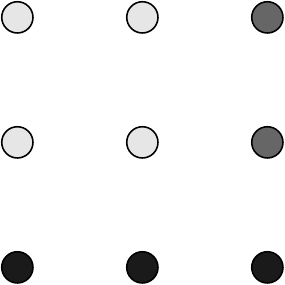}&
      \includegraphics[height=0.13\linewidth]{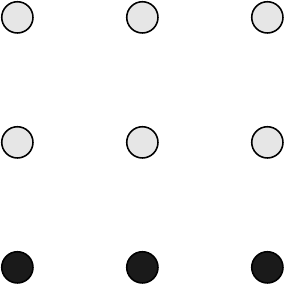}&
      \includegraphics[height=0.13\linewidth]{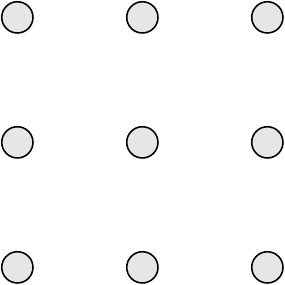}&
      \includegraphics[height=0.13\linewidth]{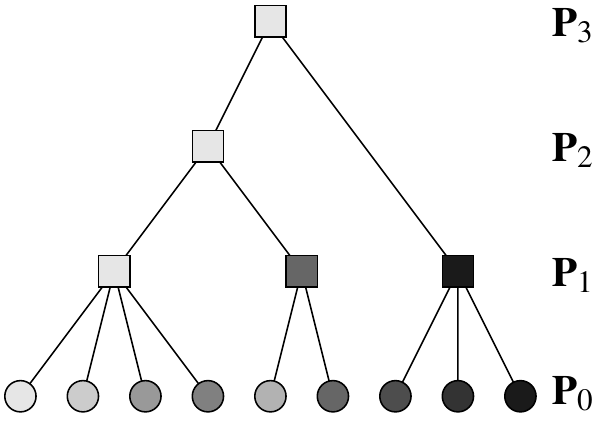}\\ 

      $\bold{P}_0$ & $\bold{P}_1$ &$\bold{P}_2$ & $\bold{P}_3$ &
      $\mathcal{H}$
\end{tabular*}
\caption{\label{fig:hierarchy} Illustration of a
  hierarchy~$\mathcal{H} =
  (\bold{P}_0,\bold{P}_1,\bold{P}_2,\bold{P}_3)$. For every partition,
  each region is represented by a gray level: two dots with the same
  gray level belong to the same region. The last subfigure represents
  the hierarchy~$\mathcal{H}$ as a tree, often called a dendrogram,
  where the inclusion relation between the regions of the successive
  partitions is represented by line segments. }
  \end{center}
\end{figure}

Figure~\ref{fig:hierarchy} graphically represents a hierarchy~$\mathcal{H} =
(\bold{P}_0,\bold{P}_1,\bold{P}_2,\bold{P}_3)$ on a rectangular
subset~$V$ of~$\mathbb{Z}^2$ made of 9 dots. For instance, it can be
seen that~$\bold{P}_1$ is a refinement of~$\bold{P}_2$ since any
region of~$\bold{P}_1$ is included in a region of~$\bold{P}_2$. It can
also be seen that the hierarchy is complete since~$\bold{P}_0$ is made
of singletons and~$\bold{P}_3$ is made of a single region that
contains all elements.

In this article, we consider connected regions, the connectivity being
given by a graph. Therefore, we remind basic graph definitions before
introducing connected partitions and hierarchies.

A {\em (undirected) graph} is a pair~$G = (V, E)$, where~$V$ is a
finite set and~$E$ is composed of unordered pairs of distinct elements
in~$V$, {\em i.e.},~$E$ is a subset of~$\left\{\{x,y\} \subseteq V \st
x \neq y \right\}$. Each element of~$V$ is called a {\em vertex or a
  point (of~$G$)}, and each element of~$E$ is called an {\em edge} (of
$G$). A {\em subgraph of $G$} is a graph $G'=(V',E')$ such that $V'$
is a subset of~$V$, and $E'$ is a subset of~$E$. If~$G'$ is a subgraph
of~$G$, we write~$G' \sqsubseteq G$. The vertex and edge sets of a
graph~$X$ are denoted by~$V(X)$ and~$E(X)$ respectively.

Let~$G$ be a graph and let~$(x_0, \ldots, x_k)$ be a sequence of
vertices of~$G$. The sequence~$(x_0, \ldots, x_k)$ is a {\em path
  (in~$G$) from~$x_0$ to~$x_k$} if, for any~$i$ in~$\{1,
\ldots,k\}$, $\{x_{i-1},x_i\}$ is an edge of~$G$. The graph~{\em
  $G$ is connected} if, for any two vertices~$x$ and~$y$ of~$G$, there
exists a path from~$x$ to~$y$. Let~$X$ be a subset of~$V(G)$. The {\em
  graph induced by~$X$ (in~$G$)} is the graph whose vertex set is~$X$
and whose edge set contains any edge of~$G$ which is made of two
elements in~$X$. If the graph induced by~$X$ is connected, we also
say, for simplicity, that~{\em $X$ is connected (for~$G$)} . The
subset~$X$ of~$V(G)$ is a {\em connected component of~$G$} if it is
connected for~$G$ and maximal for this property, {\em i.e.}, for any
subset~$Y$ of~$V(G)$, if~$Y$ is a connected superset of~$X$, then we
have~$Y = X$. In the following, we denote by~$\bold{C}(G)$ the set of
all connected components of~$G$. It is well-known that this
set~$\bold{C}(G)$ of all connected components of~$G$ is a partition
of~$V(G)$. This partition is called the {\em (connected components)
  partition induced by~$G$}. Thus, the set~$[\bold{C}(G)]_x$ is the
unique connected component of~$G$ that contains~$x$.

Given a graph~$G = (V,E)$, a {\em partition of~$V$ is connected
  (for~$G$)} if any of its regions is connected and a {\em hierarchy
  on~$V$ is connected (for~$G$)} if any of its partitions is
connected.

\begin{figure}[htb]
    \begin{center}
    \begin{tabular*}{1\linewidth}{@{\extracolsep{\fill}}c c c c c c}
      \includegraphics[height=0.15\linewidth]{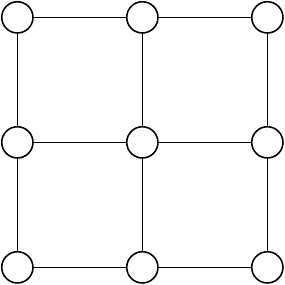}&
      \includegraphics[height=0.15\linewidth]{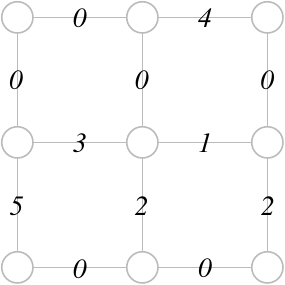}&
      \includegraphics[height=0.13\linewidth]{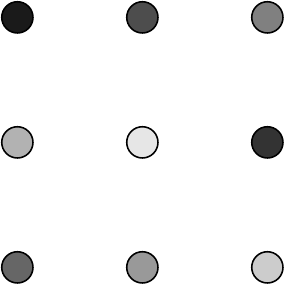}&
      \includegraphics[height=0.13\linewidth]{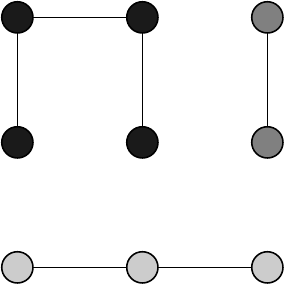}&
      \includegraphics[height=0.13\linewidth]{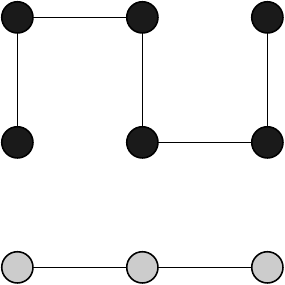}&
      \includegraphics[height=0.13\linewidth]{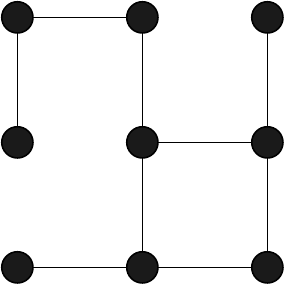}\\
      (a) & (b)&      (c) &(d) &(e) & (f)
    \end{tabular*}
\caption{\label{fig:wgraphs} Illustration of quasi-flat zones
  hierarchy. (a) A graph~$G$; (b) a map~$w$ (numbers in black) that
  weights the edges of~$G$ (in gray); (c, d, e, f) the
  $\lambda$-subgraphs of~$G$, with~$\lambda = 0,1,2,3$. The associated
  connected component partitions that form the hierarchy of
  quasi-flat zones of~$G$ for~$w$ is depicted in
  \Xfig{fig:hierarchy}.}
  \end{center}
\end{figure}

For instance, the partitions presented in \Xfig{fig:hierarchy} are
connected for the graph given in \Xfig{fig:wgraphs}(a). Therefore, the
hierarchy~$\mathcal{H}$ made of these partitions, which is depicted as
a dendrogram in \Xfig{fig:hierarchy} (rightmost subfigure), is also
connected for the graph of \Xfig{fig:wgraphs}(a).

For image analysis applications, the graph~$G$ can be obtained as a
pixel or a region adjacency graph: the vertex set of~$G$ is either the
domain of the image to be processed or the set of regions of an
initial partition of the image domain. In the latter case, the regions
can in particular be ``superpixels''. In both cases, two typical
settings for the edge set of $G$ can be considered: (1) the edges
of~$G$ are obtained from an adjacency relation between the image
pixels, such as the well known 4- or 8-adjacency relations; and (2)
the edges of~$G$ are obtained by considering, for each vertex~$x$
of~$G$, the nearest neighbors of~$x$ for a distance in a (continuous)
features space onto which the vertices of~$G$ are mapped.

\section{Hierarchies of quasi-flat zones}
\label{sec:hqfz}
As established in the next sections, a connected hierarchy can be
equivalently treated by means of an edge-weighted graph. We first
recall in this section that the level-sets of any edge-weighted graph
induce a hierarchy of quasi-flat zones. This hierarchy is widely used
in image processing \cite{Meyer-Maragos-1999}.

Let~$G$ be a graph, if $w$ is a map from the edge set of~$G$ to the
set~$\mathbb{R}^+$ of positive real numbers, then the pair~$(G,w)$ is
called an {\em (edge-)weighted graph}. If~$(G,w)$ is an edge-weighted
graph, for any edge~$u$ of~$G$, the value~$w(u)$ is called the {\em
  weight of~$u$ (for $w$)}.

{\bf Important notation.} In the sequel of this paper, we consider a
weighted graph~$(G,w)$. To shorten the notations, the vertex and edge
sets of~$G$ are denoted by~$V$ and~$E$ respectively instead of~$V(G)$
and~$E(G)$. Furthermore, we assume that the vertex set of~$G$ is
connected. Without loss of generality, we also assume that the range
of~$w$ is the set~$\mathbb{E}$ of all integers from~$0$ to~$|E|-1$
(otherwise, one could always consider an increasing one-to-one mapping
from the set~$\{w(u) \st u \in E\}$ into~$\mathbb{E}$). We also denote
by~$\mathbb{E}^\bullet$ the set~$\mathbb{E} \cup \{|E|\}$.

Let~$X$ be a subgraph of~$G$ and let~$\lambda$ be an integer
in~$\mathbb{E}^\bullet$. The {\em $\lambda$-level set of~$X$ (for~$w$)} is
the set~$w_\lambda(X)$ of all edges of~$X$ whose weight is less
than~$\lambda$:
\begin{equation}
  \label{eqn:levelSet}
 w_\lambda(X) = \{u \in E(X) \st w(u) < \lambda\}.
\end{equation}
The {\em $\lambda$-level graph of~$X$} is the
subgraph~$w^V_\lambda(X)$ of~$X$ whose edge set is the $\lambda$-level
set of~$X$ and whose vertex set is the one of~$X$:
\begin{equation}
  \label{eqn:levelGraph} 
  w^V_\lambda(X) = (V(X) , w_\lambda(X)).
\end{equation}
The connected component partition~$\bold{C}(w^V_\lambda(X)) $ induced
by the~$\lambda$-level graph of~$X$ is called the~{\em $\lambda$-level
  partition of~$X$ (for~$w$)}.

For instance, let us consider the graph~$G$ depicted in
\Xfig{fig:wgraphs}(a) and the map~$w$ shown in
\Xfig{fig:wgraphs}(b). The $0$-, $1$-, $2$- and $3$-level sets of~$G$
contain the edges depicted in \Xfigs{fig:wgraphs}(c), (d), (e), and
(f), respectively. The graphs depicted in these figures are the
associated $0$-, $1$-, $2$- and $3$-level graphs of~$G$ and the
associated $0$-, $1$-, $2$- and $3$-level partitions are shown in
\Xfig{fig:hierarchy}.

Let~$X$ be a subgraph of~$G$. If $\lambda_1$ and~$\lambda_2$ are two
elements in~$\mathbb{E}^\bullet$ such that~$\lambda_1 \leq \lambda_2$, it
can be seen that any edge of the~$\lambda_1$-level graph of~$X$ is
also an edge of the $\lambda_2$-level graph of~$X$. Thus, if two
points are connected for the~$\lambda_1$-level graph of~$X$, then they
are also connected for the~$\lambda_2$-level graph of~$X$. Therefore,
any connected component of the~$\lambda_1$-level graph of~$X$ is
included in a connected component of the~$\lambda_2$-level graph
of~$X$. In other words, the $\lambda_1$-level partition of~$X$ is a
refinement of the~$\lambda_2$-level partition of~$X$. Hence, the
sequence
\begin{equation}
  \label{eqn:flatZones}
\mathcal{QFZ}(X,w) = (\bold{C}(w^V_\lambda(X)) \st \lambda \in
\mathbb{E}^\bullet)
\end{equation}
 of all $\lambda$-level partitions of~$X$ is a hierarchy. This
 hierarchy~$\mathcal{QFZ}(X,w)$ is called the {\em quasi-flat zones
   hierarchy of~$X$ (for~$w$)}. It can be seen that this hierarchy is
 complete whenever~$X$ is connected.

For instance, the quasi-flat zones hierarchy of the graph~$G$
(\Xfig{fig:wgraphs}(a)) for the map~$w$ (\Xfig{fig:wgraphs}(b)) is the
hierarchy of \Xfig{fig:hierarchy}.

For image analysis applications, we often consider that the weight of
an edge~$u = \{x,y\}$ represents the dissimilarity of~$x$ and~$y$. For
instance, in the case where the vertices of~$G$ are the pixels of a
grayscale image, the weight~$w(u)$ can be the absolute difference of
intensity between $x$ and~$y$. The setting of the graph~$(G,w)$
depends on the application context. \ELIMINE{However, when the weights are
given by the simple gradient of intensity defined above, the connected
component partition of the~$1$-level graph of~$G$ is the partition of
the image into flat zones. Indeed the gray-level in each region of
this partition is constant and each region is a maximal connected set
satisfying this property. More generally, it can be seen that, in this
weighted graph, the definition of quasi-flat zones correspond exactly
to the usual definition for grayscale images. However, the proposed
definition can be applied to any kind of images and more generally to
edge-weighted graphs.}

\section{Correspondence between hierarchies and saliency maps}
\label{sec:sm}
In the previous section, we have seen that any edge-weighted graph
induces a connected hierarchy of partitions (called the quasi-flat
zones hierarchy). In this section, we tackle the inverse problem:
\begin{itemize}[leftmargin=0.9cm]
\item[$(P_1)$] given a connected hierarchy~$\mathcal{H}$, find a map~$w$
  from~$E$ to~$\mathbb{E}$ such that the quasi-flat zones hierarchy
  for~$w$ is precisely~$\mathcal{H}$.
\end{itemize}

We start this section by defining the saliency map
of~$\mathcal{H}$. Then, we provide a one-to-one correspondence (also
known as a bijection) between saliency maps and hierarchies. This
correspondence is given by the hierarchy of quasi flat-zones. Finally,
we deduce that the saliency map of~$\mathcal{H}$ is a solution to
problem~$(P_1)$.

Until now, we handled the regions of a partition. Let us now study
their ``dual'' that represents ``borders'' between regions and that
are called graph-cuts or simply cuts. The notion of a cut will then be
used to define the saliency maps.

Let~$\bold{P}$ be a partition of~$V$, the {\em cut of~$\bold{P}$
  (for~$G$)}, denoted by~$\phi(\bold{P})$, is the set of edges of~$G$
whose two vertices belong to different regions of~$\bold{P}$:
\begin{equation}
  \label{eqn:cut}
  \phi\left ( \bold{P} \right ) = \left \{ \left \{x,y \right \} \in
  E \st [\bold{P}]_x \neq
  [\bold{P}]_y\right \}.
\end{equation}

Let~$\mathcal{H} = (\bold{P}_0, \ldots, \bold{P}_\ell)$ be a hierarchy
on~$V$. The {\em saliency map of~$\mathcal{H}$} is the
map~$\Phi(\mathcal{H})$ from~$E$ to $\{0, \ldots, \ell\}$ such that
the weight of~$u$ for~$\Phi(\mathcal{H})$ is the maximum
value~$\lambda$ such that~$u$ belongs to the cut
of~$\bold{P}_\lambda$:
\begin{equation}
\label{eqn:saliency}
  \Phi \left (\mathcal{H} \right ) \left ( u \right ) = \max \left\{
  \lambda \in \left\{ 0,\ldots, \ell \right\} \st u \in \phi \left(
  \bold{P}_\lambda \right ) \right \}.
\end{equation}

In fact, the weight of the edge~$u = \{x,y\}$ for~$\Phi(\mathcal{H})$
is directly related to the lowest index of a partition in the
hierarchy~$\mathcal{H}$ for which~$x$ and~$y$ belong to the same
region:
\begin{equation}
  \label{eqn:saliencyLca}
  \Phi \left (\mathcal{H}\right ) \left (u\right) = \min \left \{ \lambda \in \left \{1,\ldots,
  \ell \right \} \st
  \left [\bold{P}_\lambda \right ]_x = \left [\bold{P}_\lambda \right ]_y\right \} - 1.
\end{equation}

\begin{figure}[htb]
    \begin{center}
      \begin{tabular*}{1\linewidth}{@{\extracolsep{\fill}}c c c c c c}
        \includegraphics[height=0.15\linewidth]{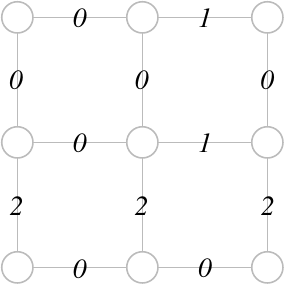}&
        \includegraphics[height=0.15\linewidth]{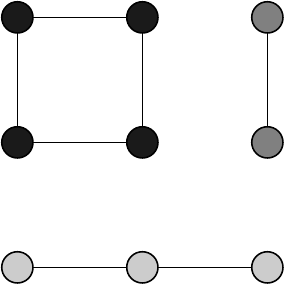}&
        \includegraphics[height=0.15\linewidth]{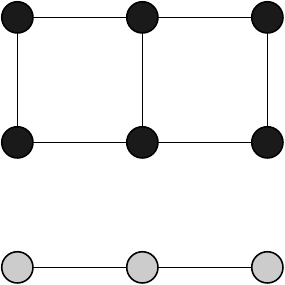}&
        \includegraphics[height=0.15\linewidth]{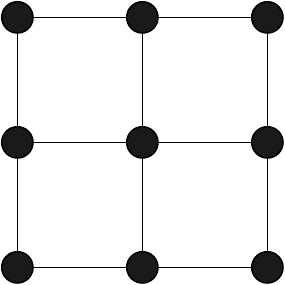}&
        \includegraphics[height=0.15\linewidth,trim=8 8 8 8,clip=true]{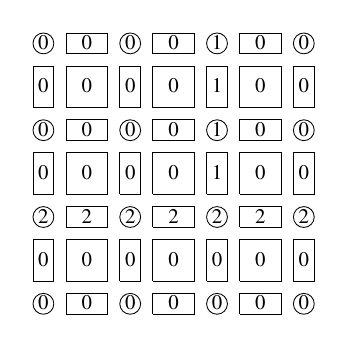}&
        \includegraphics[height=0.15\linewidth]{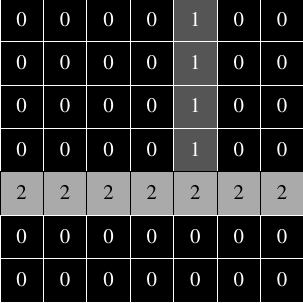}\\
        (a) & (b) & (c) & (d) & (e) & (f)
      \end{tabular*}
\caption{\label{fig:saliency} Illustration of a saliency map. The map
  (depicted by black numbers) is the saliency map~$s =
  \Phi(\mathcal{H})$ of the hierarchy~$\mathcal{H}$ shown
  in~\Xfig{fig:hierarchy} when we consider the graph~$G$ depicted in
  gray. (b, c, d) the~1-,~2-, and~3-level graphs of~$G$ for~$s$. The
  vertices are colored according to the associated 1-,~2-, and~3-level
  partitions of~$G$: in each subfigure, two vertices belonging to a
  same connected components have the same grey level. Subfigures (e)
  and (f) show possible image representations of a saliency map when
  one considers the 4-adjacency graph.}
  \end{center}
\end{figure}

For instance, if we consider the graph~$G$ represented by the gray
dots and line segments in \Xfig{fig:saliency}(a), the saliency map of
the hierarchy~$\mathcal{H}$ shown in \Xfig{fig:hierarchy} is the map
shown with black numbers in \Xfig{fig:saliency}(a). When the
4-adjacency relation is used, a saliency map can be displayed as an
image (\Xfigs{fig:saliency}(e,f) and \Xfig{fig:SaillanceSilvio}) which
is useful for visualizing the associated hierarchy at a glance.

We say that a map~$w$ from~$E$ to~$\mathbb{E}$ is a {\em saliency map}
if there exists a hierarchy~$\mathcal{H}$ such that~$w$ is the
saliency map of~$\mathcal{H}$ (\ie{} $w = \Phi(\mathcal{H})$).

If~$\varphi$ is a map from a set~$S_1$ to a set~$S_2$ and
if~$\varphi^{-1}$ is a map from~$S_2$ to~$S_1$ such that the
composition of~$\varphi^{-1}$ with~$\varphi$ is the identity, then we
say that~$\varphi^{-1}$ is the inverse of~$\varphi$.

The next theorem identifies the inverse of the map~$\Phi$ and asserts
that there is a bijection between the saliency maps and the connected
hierarchies on~$V$.

\begin{thm}
  \label{thm:SaliencyBijectionConnectedHierarchy}
  The map~$\Phi$ is a one-to-one correspondence between the connected
  hierarchies on~$V$ of depth~$|E|$ and the saliency maps (of
  range~$\mathbb{E})$. The inverse~$\Phi^{-1}$ of~$\Phi$ associates to
  any saliency map~$w$ its quasi-flat zones hierarchy: $\Phi^{-1}(w) =
  \mathcal{QFZ}(G,w)$.
\end{thm}

Hence, as a consequence of this theorem, we have~$\mathcal{QFZ}(G,
\Phi(\mathcal{H})) = \mathcal{H}$, which means that~$\mathcal{H}$ is
precisely the hierarchy of quasi-flat zones of~$G$ for its saliency
map $\Phi(\mathcal{H})$. In other words, the saliency map
of~$\mathcal{H}$ is a solution to problem~$(P_1)$. For instance, if we
consider the hierarchy~$\mathcal{H}$ shown in \Xfig{fig:hierarchy}, it
can be observed that the quasi-flat zones hierarchy
for~$\Phi(\mathcal{H})$ (see \Xfig{fig:saliency}) is
indeed~$\mathcal{H}$. We also deduce that, for any saliency map~$w$,
the relation~$\Phi(\mathcal{QFZ}(G,w)) = w$ holds true. In other words,
a given saliency map~$w$ is precisely the saliency map of its
quasi-flat zones hierarchy.

From this last relation, we can deduce that there are some maps that
weight the edges of~$G$ and that are not saliency maps. Indeed, in
general, a map is not equal to the saliency map of its quasi-flat
zones hierarchy. For instance, the map~$w$ in \Xfig{fig:wgraphs} is
not equal to the saliency map of its quasi-flat zones hierarchy which
is depicted in \Xfig{fig:saliency}. Thus, the map~$w$ is not a
saliency map. The next section studies a characterization of saliency
maps.

\section{Characterization of saliency maps}
\label{sec:saliencyCharac}
Following the conclusion of the previous section, we now consider the
problem: \ELIMINE{As seen in the previous section, given a hierarchy,
  there might well exist at least two distinct maps~$w$ and~$w'$ such
  that the quasi-flat zones hierarchy for both~$w$ and $w'$ is equal
  to~$\mathcal{H}$. Hence, in order to select among two such maps, it
  is interesting to consider the following problem:}
\begin{itemize}[leftmargin=0.9cm]
  \item[$(P_2)$] given a hierarchy~$\mathcal{H}$, find the minimal
    map~$w$ such that the quasi-flat zones hierarchy for~$w$ is
    precisely~$\mathcal{H}$. 
\end{itemize}
The next theorem establishes that the saliency map of~$\mathcal{H}$ is
the unique solution to problem~$(P_2)$.

Before stating \Xtheo{thm:SaliencyCharacterization}, let us recall
that, given two maps~$w$ and~$w'$ from~$E$ to~$\mathbb{E}$, the  map~$w'$
is less than~$w$ if we have~$w'(u) \leq w(u)$ for any~$u \in E$.

\begin{thm}
  \label{thm:SaliencyCharacterization}
  Let~$\mathcal{H}$ be a hierarchy and let~$w$ be a map from~$E$
  to~$\mathbb{E}$. The map~$w$ is the saliency map of~$\mathcal{H}$
  if and only if the two following statements hold true:
  \begin{enumerate}
  \item the quasi-flat zones hierarchies for~$w$ is~$\mathcal{H}$; and
  \item the map~$w$ is minimal for statement 1, {\em i.e.}, for any
    map~$w'$ such that~$w' \leq w$, if the quasi-flat zones hierarchy
    for~$w'$ is~$\mathcal{H}$, then we have~$w = w'$.
  \end{enumerate} 
\end{thm}

Given a weighted graph~$(G,w)$, it is sometimes interesting to
consider the saliency map of its quasi-flat zones hierarchy. This
saliency map is simply called the {\em saliency map of~$w$} and is
denoted by $\Psi(w)$. From \Xtheo{thm:SaliencyCharacterization}, the
operator~$\Psi$ which associates to any map~$w$ the saliency
map~$\Phi(\mathcal{QFZ}(G,w))$ of its quasi-flat zones hierarchy, is
idempotent (\ie{} $\Psi(\Psi(w)) =\Psi(w)$). Furthermore, it is easy to see
that~$\Psi$ is also anti-extensive (we have~$\Psi(w) \leq w$) and
increasing (for any two maps~$w$ and $w'$, if $w \geq w'$, then we
have~$\Psi(w) \geq \Psi(w')$). Thus,~$\Psi$ is a morphological
opening. This operator is studied, in different frameworks, under
several names (see
\cite{leclerc-1981,Najman-jmiv2011,Kiran-Serra-2013,Ronse-jmiv2014,nakache-confais-2004}).
\ELIMINE{: ultrametric opening \cite{leclerc-1981}, ultrametric
  watershed \cite{Najman-jmiv2011}, class opening
  \cite{Kiran-Serra-2013} or subdominant ultrametric
  \cite{nakache-confais-2004} when the complete graph is
  considered. In classification, links between subdominant ultrametric
  and single linkage clustering are well studied. It is also known
  that single linkage clustering and subdominant ultrametric can be
  recovered from the minimum spanning tree of the complete graph. The
  next section proposes a generalization of this result.} When the
considered graph~$G$ is complete, it is known in classification (see,
\eg{}, \cite{nakache-confais-2004}) that this operator is linked to
the minimum spanning tree of~$(G,w)$. The next section proposes a
generalization of this link.

\section{Minimum spanning trees}
\label{sec:mst}
Two distinct maps that weight the edges of the same graph (see, \eg{},
the maps of \Xfigs{fig:wgraphs}(b) and \ref{fig:saliency}(a)) can
induce the same hierarchy of quasi-flat zones. Therefore, in this
case, one can guess that some of the edge weights do not convey any
useful information with respect to the associated quasi-flat zones
hierarchy. More generally, in order to represent a hierarchy by a
simple (\ie{}, easy to handle) edge-weighted graph with a low level
of redundancy, it is interesting to consider the following problem:
\begin{itemize}[leftmargin=0.9cm] 
\item[$(P_3)$] given an edge-weighted graph $(G,w)$, find a minimal
  subgraph~$X \sqsubseteq G$ such that the quasi-flat zones
  hierarchies of~$G$ and of~$X$ are the same.
\end{itemize}
The main result of this section, namely \Xtheo{thm:SaliencyMST},
provides the set of all solutions to problem $(P_3)$: the minimum
spanning trees of~$(G,w)$. The minimum spanning tree problem is
one of the most typical and well-known problems of combinatorial
optimization (see \cite{cormen-et-al-2001}) and
\Xtheo{thm:SaliencyMST} provides, as far as we know, a new
characterization of minimum spanning trees based on the quasi-flat
zones hierarchies as used in image processing.

Let~$X$ be a subgraph of~$G$. The weight of~$X$ with respect to~$w$,
denoted by~$w(X)$, is the sum of the weights of all the edges
in~$E(X)$:~$w(X) = \sum_{u \in E(X)} w(u)$.  The subgraph~$X$ is a
{\em minimum spanning tree (MST) of~$(G,w)$} if:
\begin{enumerate}
\item $X$ is connected; and
\item $V(X) = V$; and
\item the weight of~$X$ is less than or equal to the weight of any
  graph~$Y$ satisfying (1) and (2) ({\em i.e.},~$Y$ is a connected
  subgraph of~$G$ whose vertex set is~$V$).
\end{enumerate}

\begin{figure}[htb]
    \begin{center}
    \begin{tabular*}{1\linewidth}{@{\extracolsep{\fill}}c c c c}
      \includegraphics[height=0.15\linewidth]{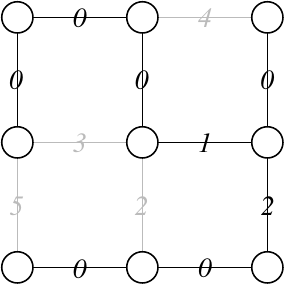}&
      \includegraphics[height=0.15\linewidth]{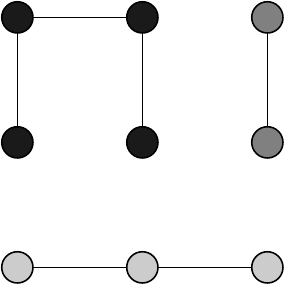}&
      \includegraphics[height=0.15\linewidth]{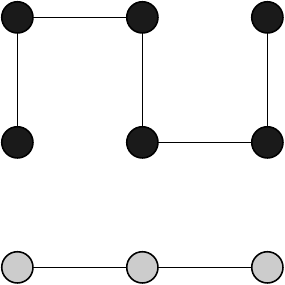}&
      \includegraphics[height=0.15\linewidth]{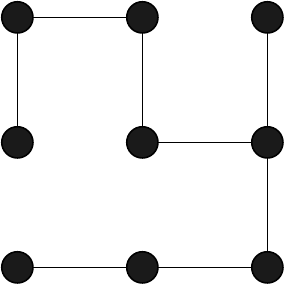}\\ 
      (a) & (b)   &(c)  & (d)  
      
\end{tabular*}
\caption{\label{fig:MST} Illustration of a minimum spanning tree and
  of its quasi-flat zones hierarchy. (a) A minimum spanning tree~$X$
  (black edges and black circled vertices) of the weighted graph
  of~\Xfig{fig:wgraphs}(b); (b, c, d) the~1- ,~2-, and~3-level graphs
  of~$X$. The vertices are colored according to the associated 1-,~2-,
  and~3-level partitions of~$X$: in each subfigure, two vertices
  belonging to the same connected components have the same color. }
  \end{center}
\end{figure}

For instance, a MST of the graph shown in \Xfig{fig:wgraphs}(b) is
presented in \Xfig{fig:MST}(a).

\ELIMINE{Note that there can be several distinct MSTs of~$(G,w)$, but
  that any MST of~$(G,w)$ is a {\em tree spanning~$V$}, {\em i.e.}, a
  connected subgraph of~$G$ with~$|V|$ vertices and with~$|V|-1$
  edges.}

\begin{thm}
  \label{thm:SaliencyMST}
  A subgraph~$X$ of~$G$ is a MST of~$(G,w)$ if and only if the two
  following statements hold true:
  \begin{enumerate}
    \item the quasi-flat zones hierarchies of~$X$ and of~$G$ are the
      same; and
    \item the graph~$X$ is minimal for statement 1, {\em i.e.}, for
      any subgraph~$Y$ of~$X$, if the quasi-flat zones hierarchy
      of~$Y$ for~$w$ is the one of~$G$ for~$w$, then we have~$Y = X$.
      \end{enumerate}
\end{thm}

Theorem~\ref{thm:SaliencyMST} (statement 1) indicates that the
quasi-flat zones hierarchy of a graph and of its MSTs are
identical. Note that statement 1 appeared in
\cite{Cousty-Perret-Najman-ismm2013} but \Xtheo{thm:SaliencyMST}
completes the result of \cite{Cousty-Perret-Najman-ismm2013}. Indeed,
\Xtheo{thm:SaliencyMST} indicates that there is no proper subgraph of
a MST that induces the same quasi-flat zones hierarchy as the initial
weighted graph. Thus, a MST of the initial graph is a solution to
problem $(P_3)$, providing a minimal graph representation of the
quasi-flat zones hierarchy of~$(G,w)$, or more generally by
\Xtheo{thm:SaliencyBijectionConnectedHierarchy} of any connected
hierarchy. More remarkably, the converse is also true: a minimal
representation of a hierarchy in the sense of~$(P_3)$ is necessarily a
MST of the original graph. To the best of our knowledge, this result
has not been stated before.

For instance, the level sets, level graphs and level partitions of the
MST~$X$ (\Xfig{fig:MST}(a)) of the weighted graph~$(G,w)$
(\Xfig{fig:wgraphs}) are depicted in \Xfigs{fig:MST}(b), (c), (d). It
can be observed that the level partitions of~$X$ are indeed the same
as those of~$G$. Thus the quasi-flat zones hierarchies of~$X$ and~$G$
are the same.

\section{Saliency map algorithms}
When a hierarchy~$\mathcal{H}$ is stored as a tree data structure
(such as \eg{} the dendrogram of \Xfig{fig:hierarchy}), the weight of
any edge for the saliency map of this hierarchy \ELIMINE{(also known
  as the connection value
  \cite{Bertrand-2005,Couprie-et-al-2005,Cousty-et-al-2010} between
  the two extremities of the edge)} can be computed in constant time,
provided a linear time preprocessing. Indeed, the weight of an edge
linking~$x$ and~$y$ is associated (see \Xeqn{eqn:saliencyLca}) to the
lowest index of a partition for which~$x$ and~$y$ belongs to the same
region. This index can be obtained by finding the index of the least
common ancestor of~$\{x\}$ and~$\{y\}$ in the tree. The algorithm
proposed in \cite{Bender-Farach-Colton-2000} performs this task in
constant time, provided a linear time preprocessing of the
tree. Therefore, computing the saliency map of~$\mathcal{H}$ can be
done in linear~$O(|E| +|V|)$ time complexity.

Thus, the complete process that computes the saliency map~$\Psi(w)$ of
a given map~$w$ proceeds in two steps:
\begin{itemize}
\item[i)] build the quasi-flat zones hierarchy~$\mathcal{H} =
  \mathcal{QFZ}(G,w)$ of~$G$ for~$w$; and
\item[ii)] compute the saliency map~$\Psi(w) = \Phi(\mathcal{H})$.
\end{itemize}
On the basis of \cite{Cousty-Perret-Najman-ismm2013}, step i) can be
performed with the quasi-linear time algorithm shown in \cite{Najman-et-al-ismm2013}
and step ii) can be performed in linear-time as proposed in the
previous paragraph. Thus, the overall time complexity of this
algorithm is quasi-linear with respect to the size~$|E| + |V|$ of the
graph~$G$.

The algorithm sketched in \cite{Najman-jmiv2011}, based on
\cite{Couprie-et-al-2005}, for computing the saliency map of a given
map~$w$ has the same complexity as the algorithm proposed
above. However, the algorithm of \cite{Najman-jmiv2011} is more
complicated since it requires to compute the topological watershed of
the map. This involves a component tree (a data structure which is
more complicated than the quasi-flat zones hierarchy in the sense of
\cite{Cousty-Perret-Najman-ismm2013}), a structure for computing least
common ancestors, and a hierarchical queue \cite{Couprie-et-al-2005},
which is not needed by the above algorithm. Hence, as far as we know,
the algorithm presented in this section is the simplest algorithm for
computing a saliency map. It is also the most efficient both from
memory and execution-time points of view. An implementation in C of
this algorithm is available at \url{http://www.esiee.fr/\~{}info/sm}.

\section{Conclusions}
In this article we study three representations for a hierarchy of
partitions. We show a new bijection between hierarchies and saliency
maps and we characterize the saliency map of a hierarchy and the
minimum spanning trees of a graph as minimal elements preserving
quasi-flat zones. In practice, these results allow us to indifferently
handle a hierarchy by a dendrogram (the direct tree structure given by
the hierarchy), by a saliency map, or by an edge-weighted tree. These
representations form a toolkit for the design of hierarchical
(segmentation) methods where one can choose the most convenient
representation or the one that leads to the most efficient
implementation for a given particular operation. The results of this
paper were used in \cite{Guimaraes-et-al-ssspr2012} to provide a
framework for hierarchicalizing a certain class of non-hierarchical
methods. We study in particular a hierarchicalization of
\cite{Felzenszwalb-ijcv2004}. The first results are encouraging and a
short term perspective is the precise practical evaluation of the gain
of the hierarchical method with respect to its non-hierarchical
counterpart.

Another important aspect of the present work is to underline and to
precise the close link that exists between classification and
hierarchical image segmentation. Whereas classification methods were
used as image segmentation tools for a long time, our results incite
us to look if some hierarchies initially designed for image
segmentation can improve the processing of non-image data such as data
coming from geography, social network, etc.. This topic will be a
subject of future research.

\ELIMINE{
\Xtheos{thm:SaliencyBijectionConnectedHierarchy}
and~\ref{thm:SaliencyMST} indicate that any connected hierarchy can be
handled by means of a weighted spanning tree which is a subgraph
of~$G$.\ELIMINE{ Indeed, by
  \Xtheo{thm:SaliencyBijectionConnectedHierarchy}, we can associate to
  any hierarchy~$\mathcal{H}$ the weight map~$\Phi(\mathcal{H})$ whose
  quasi-flat zones hierarchy is~$\mathcal{H}$. Then, we can restrict
  the function~$\Phi(\mathcal{H})$ to a MST~$X$
  of~$\Phi(\mathcal{H})$, whose quasi-flat zones hierarchy is still,
  by \Xtheo{thm:SaliencyMST}, the hierarchy~$\mathcal{H}$. Thus, any
  hierarchy can be represented by a weighted spanning tree which
  requires less memory than the original graph and is easier to handle
  because it contains less redundancy. }This representation of
hierarchies is used in \cite{Guimaraes-et-al-ssspr2012} for proposing
an algorithm that transforms a first hierarchy into another based on
some features of the first hierarchy. Based on this representation, an
``elementary modification'' of the hierarchy is made by changing the
weight of single edge in the MST. The result of the method presented
in \cite{Guimaraes-et-al-ssspr2012} is presented in
\Xfig{fig:SaillanceSilvio} in the form of a saliency map. Future work
include the experimental assessment of this method.

Another stimulating perspective suggested by the links drawn in this
paper with hierarchical classification is the application of
morphological hierarchies to non-necessarily image data.
}
\bibliographystyle{splncs03}
\bibliography{cousty_bibtexfile}

\end{document}